\documentclass{ifacconf}
\makeatletter
\let\old@ssect\@ssect
\makeatother

\usepackage{hyperref}

\makeatletter
\def\@ssect#1#2#3#4#5#6{%
  \NR@gettitle{#6}% 
  \old@ssect{#1}{#2}{#3}{#4}{#5}{#6}% 
}
\makeatother

\usepackage{graphicx}
\usepackage{amsmath}
\usepackage{float}
\usepackage{pifont}
\usepackage{pgf}
\usepackage{tikz}
\usepackage{commath}
\usepackage{epstopdf}
\usepackage{ifthen}
\usepackage{bbm}
\usepackage{dsfont}
\usepackage{amssymb}
\usepackage{txfonts}
\usepackage{mathbbol}
\usepackage{multirow}
\usepackage{lipsum}
\usepackage[linesnumbered, ruled, vlined, algo2e]{algorithm2e}
\usepackage{makecell}
\usepackage{natbib}
\SetKwInput{KwInput}{Input}
\SetKwInput{KwOutput}{Output}

\begin{document}
\begin{frontmatter}
\title{Robust Deep Reinforcement Learning for Quadcopter Control}
\author{Aditya M. Deshpande, Ali A. Minai, Manish Kumar}
\address{University of Cincinnati, 2600 Clifton Ave., Cincinnati, Ohio 45221\\
(e-mail: deshpaad@mail.uc.edu, ali.minai@uc.edu, manish.kumar@uc.edu)}

\begin{abstract}
Deep reinforcement learning (RL) has made it possible to solve complex robotics problems using neural networks as function approximators. However, the policies trained on stationary environments suffer in terms of generalization when transferred from one environment to another. In this work, we use Robust Markov Decision Processes (RMDP) to train the drone control policy, which combines ideas from Robust Control and RL. It opts for pessimistic optimization to handle potential gaps between policy transfer from one environment to another. The trained control policy is tested on the task of quadcopter positional control. RL agents were trained in a MuJoCo simulator. During testing, different environment parameters (unseen during the training) were used to validate the robustness of the trained policy for transfer from one environment to another. The robust policy outperformed the standard agents in these environments, suggesting that the added robustness increases generality and can adapt to nonstationary environments.
\\
\\
Codes: \textcolor{blue}{\textbf{\url{https://github.com/adipandas/gym_multirotor}}}

\end{abstract}

\begin{keyword}
Reinforcement learning control; Robust adaptive control; Robotics; Flying robots
\end{keyword}

\end{frontmatter}

\section{Introduction}
\vspace{-5pt}
Unmanned aerial vehicles (UAVs) and their applications have shown an unprecedented increase in recent times. To name a few, drones have been used for surveillance and inspection [\cite{bethke2008group}, \cite{gonzalez2016unmanned}, \cite{gonccalves2015uav}], agriculture [\cite{bethke2008group}], package delivery [\cite{magsino2020achieving}], wildlife monitoring [\cite{hodgson2016precision}] and search and rescue [\cite{eid2019design}]. Furthermore, there have been continuous enhancements in quadcopter design with additional degrees of freedom such as tilt-rotor quadcopter design by \cite{kumar2020quaternion}, morphing quadcopter design by \cite{falanga2018foldable}, and the sliding arm drone by \cite{kumar2020flight} for dealing with various complex scenarios that may arise for different missions.

\begin{figure}[t!]
	\centering
	\includegraphics[width=\linewidth]{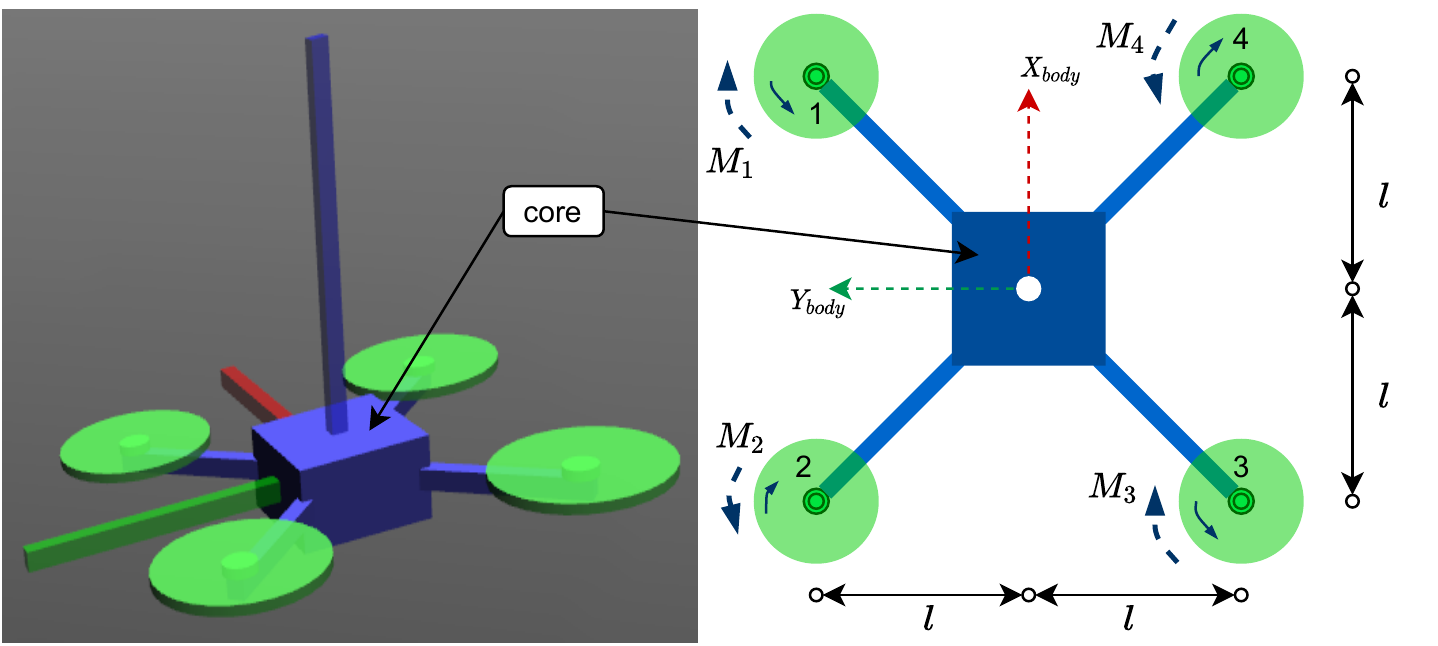}
	\vspace{-10pt}
    \caption{Quadcopter schematic. (left) $RGB$-colored body-axes on mujoco model correspond to $XYZ$-axes in body frame respectively; (right) Top-down view of generalized quadcopter with X-configuration design.}
    \label{fig:quadrotor-x-mujoco}
\end{figure}

To enable these applications, the development of control software for UAVs, especially quadcopters, has been an active area of research [\cite{santoso2017state}, \cite{meier2015px4}]. The advent of deep learning and reinforcement learning has enabled development of neural controllers for these systems [\cite{koch2019reinforcement}, \cite{hwangbo2017control}]. The neural policies are able to control the complex flying systems [\cite{adi2020devrlquad}] as well as work with high-dimensional observation spaces [\cite{falanga2020dynamic}].

Although Deep RL has found wide applications for solving complex tasks in the UAV community, the training and adaptation of a policy for different environments remains challenging. Policies learned in stationary environments using RL are fragile and tend to fail with the change in environmental parameters such as mass, moment of inertia of the system, drag, or friction in the environment. Hence, the transfer of RL policies from one environment to another requires special attention [\cite{zhao2020sim}]. \cite{molchanov2019sim} and \cite{polvara2020sim} used domain randomization techniques to learn stabilizing control policies for quadcopters. Domain randomization for RL requires maintaining distributions over the model parameters and/or perturbations which may vary across environments. This makes the policy robust only to scenarios that are encompassed by these distributions. Furthermore, these sets of distributions may not be always tractable and the range of uncertainties in which a policy would operate safely remains unknown. 
Therefore, the policies trained using domain randomization remain limited to within-the-distribution scenarios.

\begin{figure*}[ht!]
	\centering
	\includegraphics[width=0.85\textwidth]{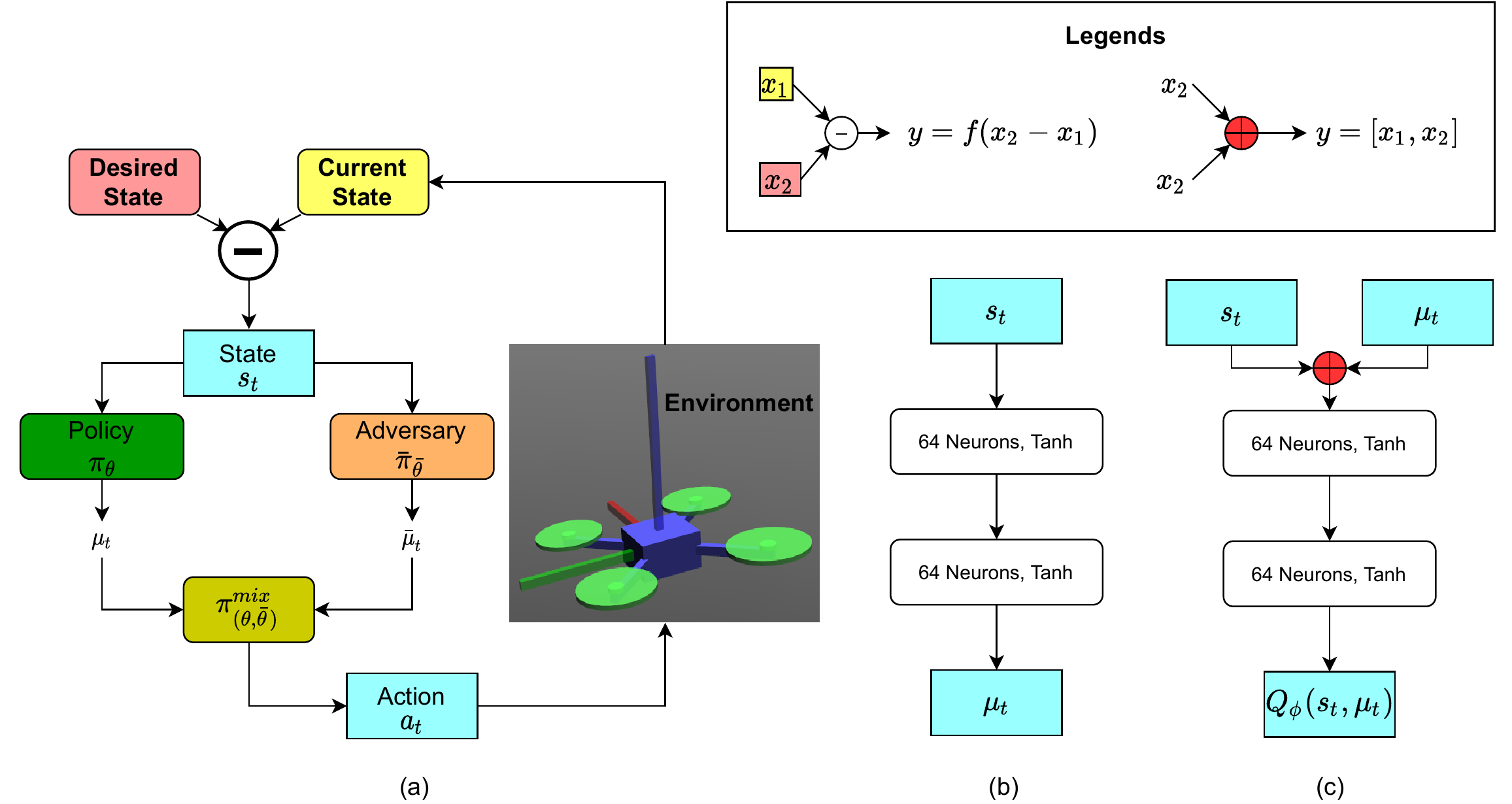}
	\vspace{-10pt}
    \caption{Neural Network Architectures: (a) The network setup used for training robust policy where $\mu_{t}$ is the action by policy $\pi_{\theta}$ and $\bar{\mu}$ is the action by the adversary $\bar{\pi}_{\bar{\theta}}$; (b) Architecture of the neural networks used for policy $\pi_{\theta}$ and for adversary $\bar{\pi}_{\bar{\theta}}$; (c) Architecture of the neural network used as the critic (action-value function) $Q_{\phi}(s_{t}, \mu_{t})$.}
    \label{fig:network-arch}
\end{figure*}

Even though domain randomization helps in generalization, most of the reinforcement learning data is sampled from simulated robot models. Due to the limitations of modeling environments, there are inconsistencies in the data from simulation and real-world even for the same task and initial conditions. Thus, the RL agents used for robot control would always experience some model mismatch and the trained agent would not perform optimally on the real robot [\cite{RobotRL2021FinnLevine}].

In controls engineering, robustness refers to the controller performance in case of environment and/or parametric uncertainties [\cite{lee2013nonlinear}]. Along similar lines, for RL, the challenge of training a robust controller is addressed using Robust Markov Decision Processes [\cite{iyengar2005robust}]. In this, the RL objective is achieved using max-min optimization over a set of possible environment uncertainties. Contrary to the prior knowledge of domain uncertainties required for domain randomization approaches, the RMDP framework enables training a policy to maximize the reward with respect to the worst possible model of the environment. We adopt the robust deep reinforcement learning approach proposed by \cite{tessler2019action} to train a quadcopter control policy in continuous state and action spaces.
Our results show the efficacy of this method by training a robust policy for a quadcopter waypoint navigation task considering the presence of both environment as well as parametric uncertainties.

\section*{Contributions}\label{sec:contributions}
\vspace{-8pt}
In this article, we study the application of robust reinforcement learning to learn a robust control policy for a quadcopter. The controller is trained using the Robust Markov Decision Process (RMDP) framework. Our results demonstrate the adaptability of the robust learned policy to external and internal system uncertainties at test time. \emph{It should be noted that the primary contribution of this work lies in the novel
application of Robust-RL for learning quadcopter controller to enable functioning in non-stationary (uncertain) environmental conditions and not in the algorithm itself.}

The remainder of this article is organized as follows: Section-\ref{sec:uav-dynamics} briefly describes the quadcopter dynamics, Section-\ref{sec:robust-rl} provides the detailed approach used for learning the robust quadcopter controller and Section-\ref{sec:env-setup} provides the details about the setup used for training the control policies. In Section-\ref{sec:results}, we discuss our results. Finally, we conclude the article and provide future directions for research in Section-\ref{sec:conclusions}.

\section{Quadcopter Dynamics Model}
\label{sec:uav-dynamics}
\vspace{-5pt}
In this section, we briefly discuss the dynamics of the quadcopter. A quadcopter of `$\mathbf{X}$'-configuration is used in the work. Figure \ref{fig:quadrotor-x-mujoco} shows the schematic diagram of the quadcopter along with its physics engine simulated model. The body frame origin of the drone is placed at its center of mass. Thrust motors of this system rotate only in one direction and produce a positive thrust along the $z_{body}$ axis.
The translational motion of the UAV in the world frame and rotational motion in the body frame is represented by equations \eqref{eq:newton_law_translation} and \eqref{eq:newton_law_rotation}, respectively.
\begin{align}
m 
\begin{pmatrix} \ddot x \\ \ddot y \\ \ddot z \end{pmatrix}
&=
\begin{pmatrix}  0 \\ 0 \\ -mg \end{pmatrix}
+ 
R 
\begin{pmatrix}  0 \\ 0 \\ \sum_{i=1}^{4} F_{i} \end{pmatrix}  \label{eq:newton_law_translation}
\\
I 
\begin{pmatrix} \dot p \\ \dot q \\ \dot r \end{pmatrix} 
&= 
\begin{pmatrix} 
    l(  F_1 + F_2 - F_3 - F_4) \\
    l(- F_1 + F_2 + F_3 - F_4) \\ 
    - M_1 + M_2 - M_3 + M_4
\end{pmatrix}
-
\begin{pmatrix} p \\ q \\ r \end{pmatrix}
\times
I 
\begin{pmatrix}  p \\ q \\ r \end{pmatrix} \label{eq:newton_law_rotation}
\end{align}
\noindent where $R\in SO(3)$ is the rotational matrix from body to world frame of reference and is given by \eqref{eq:rotation_matrix_w2b}.
\begin{align}
    R &= 
    \begin{bmatrix} 
    c_{\psi} c_{\rho} & s_\xi s_\rho c_\psi - s_\psi c_\xi & s_\xi s_\psi + s_\rho c_\xi c_\psi
    \\
    s_{\psi} c_\rho   & s_\xi s_\psi s_\rho + c_\xi c_\psi & -s_\xi c_\psi + s_\psi s_\rho c_\xi 
    \\
    -s_\rho  & s_\xi c_\rho  & c_\xi c_\rho
    \end{bmatrix}
    \label{eq:rotation_matrix_w2b}
\end{align}
In equation \eqref{eq:rotation_matrix_w2b}, the sine and cosine angle terms are represented as $s_{a}$ and $c_{a}$, respectively where $a$ represents the roll $\xi$, pitch $\rho$ and yaw $\psi$ angles.
The input thrust forces produced by each rotor are denoted by $F_i$, $\forall i \in {1,2,3,4}$. The mass of the system is represented by $m$ and the acceleration due to gravity is represented by $g$. The acceleration of the system in world frame is denoted by $\ddot{x}$, $\ddot{y}$, $\ddot{z}$, and $I$ is a diagonal matrix containing terms of moments of inertia about $x_{body}-y_{body}-z_{body}$-axes. The quadcopter length is $2 \times l$, and $p$, $q$ and $r$ are the roll-, pitch- and yaw-rates in the body frame, respectively. The torque produced by each propeller is given by $M_i$, $\forall{i}\in\lbrace 1,2,3,4\rbrace$. The thrust force and the torque produced by each motor are directly proportional to the square of the angular speed of the propeller with the constants of proportionality given by $k_f$ and $k_m$, respectively [\cite{grasp}].

\section{Training Robust control Policy}
\label{sec:robust-rl}
\vspace{-5pt}
In this paper, we consider the case of continuous state space and a continuous action space for robots. A Markov decision process (MDP) is used to model an RL problem, and it is described by the tuple of elements $(S, A, P, R, \gamma)$ where $S$ is the state space, $A$ is the action space, $P(s_{t+1}|s_{t}, a_{t})$ is the probability distribution over state transitions from $s_{t} \in S$ to $s_{t+1} \in S$ by taking action $a_{t} \in A$, $R(s_{t}, a_{t}, s_{t+1})$ is the reward function, and $0 \le \gamma < 1$ is the discount factor [\cite{sutton2018reinforcement}]. The objective of RL is to learn a control policy $\pi(a_{t}\mid s_{t})$ to maximize the expected cumulative reward in \eqref{eq:policy-obj}, where $\tau = (s_0, a_0, s_1, a_1, \dots)$ is the trajectory sampled using policy $\pi$ and the initial state is sampled from a fixed distribution $s_0 \sim p_{s_{0}}$.
\begin{align}
\begin{split}
\pi^{*} &= \arg\max_{\pi} \mathbb{E}_{\tau \sim \pi, p_{s_{0}}} \begin{bmatrix} R(\tau) \end{bmatrix} 
\\
&=
\arg\max_{\pi} \mathbb{E}_{\tau \sim \pi, p_{s_{0}}} \begin{bmatrix} \sum_{t=0}^{\infty} \gamma ^ t r_t \mid \pi \end{bmatrix} 
\end{split}
\label{eq:policy-obj}
\end{align}
Unlike RL, for Robust Reinforcement Learning, we use a Robust Markov Decision Process (RMDP) to formulate the learning objective [\cite{nilim2005robust}, \cite{iyengar2005robust}]. Mathematically, an RMDP is also modeled with a tuple of $(S, A, P, R, \gamma)$, similar to an MDP, with the only exception being that the model uncertainty is incorporated in this setup by considering the transition probabilities $P$ to be unknown but bounded in some prior known uncertainty set, i.e., $P \! \subseteq \! \mathcal{P}$. Model uncertainty is described by the set of transition models $\mathcal{P}$.

\begin{algorithm2e}
\DontPrintSemicolon
  \KwInput{
  Actor (a.k.a. Policy) training steps $n$, uncertainty value $\alpha$ and discount factor $\gamma$, maximum number of episodes $N$, maximum length of each episode $T$, learning rate $\eta$
  }
  
  Randomly initialize actor, adversary and critic networks as $\pi_{\theta}$,  $\bar{\pi}_{\bar{\theta}}$ and $Q_{\phi}(s, a)$ respectively.
  
  Initialize corresponding target network weights as: $\quad$ $\theta^{target} \leftarrow \theta$, $\bar{\theta}^{target} \leftarrow \bar{\theta}$ and $\phi^{target} \leftarrow \phi$.
  
  Initialize replay buffer $\mathcal{R}$
  
  \For{$k=\{0, 1, \dots, N\}$}    
        { Initial state $s_0 \sim p_{s_{0}}$. \;
          
          \For {$t=\{0, \dots, T\}$}
          {
          Action $a_{t}=\begin{cases} 
          \pi(s_{t}),& \text{with probability } (1-\alpha)\\
          \bar{\pi}(s_{t}), & \text{otherwise}
          \end{cases}$
          
          $\Tilde{a}_t = a_{t} + \text{exploration noise}$
          
          Execute $\Tilde{a}_t$, observe next state $s_{t+1}$, reward $r_{t}$ and episode termination flag $d$.
          
          Dump $(s_{t}, a_{t}, s_{t+1}, r_{t}, d)$ in $\mathcal{R}$.
          
          \For {$i=\{0, \dots, n\}$}{
                Sample batch $B=\{(s, a, s^{\prime}, r, d)\}$ from $\mathcal{R}$.
            
                Update $\theta$ by one step gradient ascent using 
                $\nabla_{\theta} \frac{1}{\lvert B \rvert} (1-\alpha) \sum\limits_{s \in B} Q_{\phi}(s, \pi_{\theta}(s))$.
                
                Compute targets:
                \qquad\qquad\qquad\qquad\qquad\qquad
                $\Tilde{Q} =
                    (1\!-\!\alpha) Q_{\phi^{target}}\left(s^{\prime}, \pi_{\theta^{target}}(s^{\prime})\right)
                    +
                    \alpha Q_{\phi^{target}}\left(s^{\prime}, \bar{\pi}_{\bar{\theta}^{target}}(s^{\prime})\right)$;
                    ${
                    y(r, s^{\prime}, d) = r \!+\! \gamma (1-d) \Tilde{Q}
                }$
                    
                Update $\phi$ by one step gradient descent using
                $
                \nabla_{\phi} 
                \frac{1}{\lvert B \rvert} 
                \sum\limits_{(s, a, s^{\prime}, r, d) \in B} 
                \left(
                Q_{\phi}(s, a)-y(r, s^{\prime}, d)
                \right)
                $
          }
          
          Sample batch $B=\{(s, a, s^{\prime}, r, d)\}$ from $\mathcal{R}$.
          
          Update $\bar{\theta}$ by one step gradient descent using
          $\nabla_{\bar{\theta}} \frac{1}{\lvert B \rvert} \alpha Q_{\phi}(s, \bar{\pi}_{\bar{\theta}}(s))$\\
          
          Compute targets:
          \qquad\qquad\qquad\qquad\qquad\qquad
                $\Tilde{Q} =
                    (1\!-\!\alpha) Q_{\phi^{target}}\left(s^{\prime}, \pi_{\theta^{target}}(s^{\prime})\right)
                    +
                    \alpha Q_{\phi^{target}}\left(s^{\prime}, \bar{\pi}_{\bar{\theta}^{target}}(s^{\prime})\right)$;
                    \qquad\qquad
                    ${
                    y(r, s^{\prime}, d) = r \!+\! \gamma (1-d) \Tilde{Q}
                }$
                    
            Update $\phi$ by one step gradient descent using
                $
                \nabla_{\phi} 
                \frac{1}{\lvert B \rvert} 
                \sum\limits_{(s, a, s^{\prime}, r, d) \in B} 
                \left(
                Q_{\phi}(s, a)-y(r, s^{\prime}, d)
                \right)
                $
            
          Update target networks: \qquad
            $\theta^{target}  \leftarrow \tau \theta + (1-\tau) \theta^{target}$
            $\bar{\theta}^{target}  \leftarrow \tau \bar{\theta} + (1-\tau) \bar{\theta}^{target}$
            $\phi^{target}  \leftarrow \tau \phi + (1-\tau) \phi^{target}$\\
          }
        }
\caption{Action Robust Deep Deterministic Policy Gradient Pseudo code}\label{alg:AR-DDPG}
\end{algorithm2e}

The RMDP version of the RL is an extension of the MDP by the addition of an adversarial agent. The RL agent selects actions $a_{t} \in A$ at each time step and aims to maximize returns. But another agent, with an adversarial nature, selects transition models from $\mathcal{P}$ each time step with an aim to minimize returns resulting in a sequential two-player zero-sum game. The resultant policy obtained in this game is given by equation \eqref{eq:min_max_obj}.
\begin{align}
\pi^{*}
&= 
\arg\max_{\pi} 
\min_{ \bar{\pi} } 
\mathbb{E}_{\tau \sim \pi, \bar{\pi}, p_{s_{0}}} 
\begin{bmatrix} R(\tau) \end{bmatrix} 
\label{eq:min_max_obj}
\end{align}
where $\pi^{*}$ represents the optimal control policy. Note, here we select the optimal policy as the maximizing policy, i.e., $\pi^{*} = \pi$. The policies $\pi^{*}$ (maximizer) and $\bar{\pi}^{*}$ (minimizer or adversary) learned after this optimization are said to be in Nash Equilibrium and neither of them may improve the game's outcome further.

To learn robust drone policies, we consider the robustness over the action uncertainty in drones. The actions correspond to thrust forces generated by the quadcopter. We use the Action Robust Deep Deterministic Policy Gradient (AR-DDPG) algorithm by \cite{tessler2019action} to train the robust policies. This approach requires only a scalar value  $\alpha$ to describe the influence of the adversary on the action-outcome as presented in Algorithm \ref{alg:AR-DDPG}.
The parameter $\alpha$ implicitly defines an uncertainty set $\mathcal{P}$ for the model. Thus, we do not have to provide the uncertainty set explicitly while training the policy. Here, while training the control policy $\pi_{\theta}$, the adversary $\bar{\pi}_{\bar{\theta}}$ takes the control and acts on the system with probability $\alpha$ (step $7$ in Algorithm \ref{alg:AR-DDPG}). In the case of the drone, this adversarial attack can be thought of as noise or external disturbances such as wind or model uncertainties such as errors in system identification. 
Thus, when the controller takes some action, the adversary makes the system act differently than expected. Thus, the optimization strategy of AR-DDPG used here enables the training of a robust optimal policy with respect to uncertain scenarios. The objective of the work reported here is to demonstrate the application of robust RL for training drone control policies. Readers are referred to \cite{tessler2019action} and \cite{Lillicrap2016ContinuousCW} for further mathematical elaboration of the action-robust MDPs and deep deterministic policy gradients. 

We trained the control policy in an off-policy manner. Although the adversary $\bar{\pi}$ induces a stochastic setting for training the policy (step $7$ in Algorithm \ref{alg:AR-DDPG}), noise was introduced into the action space to encourage exploration (step $8$ in Algorithm \ref{alg:AR-DDPG}). We use time-correlated Ornstein–Uhlenbeck noise for exploration as recommended by \cite{Lillicrap2016ContinuousCW}. Furthermore, to improve exploration at the start of training, we allow babbling of the quadcopter for $500$ episodes where, in each episode, the environment state is initialized randomly as described in \ref{sec:training-strategy}, and the actions are sampled from a uniform distribution in the range $[-1, 1]$ and scaled appropriately before being applied to the system (Eq. \eqref{eq:action-scaling}).

The neural networks were trained in the actor-critic style. Figure \ref{fig:network-arch} shows the setup and architectures of the neural networks used for learning the policy for the quadcopter. Figure \ref{fig:network-arch}-(a) illustrates the policy setup used during the training where 
\begin{equation}
    \pi^{mix}_{\theta, \bar{\theta}} = \begin{cases} 
          \pi(s_{t}),& \text{with probability } (1-\alpha)\\
          \bar{\pi}(s_{t}), & \text{otherwise}
          \end{cases}
\end{equation}
The architectures of the control policy and the adversary are identical and shown in Fig. \ref{fig:network-arch}-(b) and the critic or action-value function $Q(s, \mu)$ architecture is shown in Fig. \ref{fig:network-arch}-(c).
We did not apply any neural architecture optimization and did not consider the effects of variation in activation functions in the present work.
The state or observation space used for learning quadcopter control is given by the tuple $s_{t}=(e_{p_{t}}, e_{v_{t}}, R_{t}, e_{\omega_{t}})$. The control policy maps the current state of the system to the actuator commands represented by the vector $a_{t}=\{F_{1, t}, F_{2, t}, F_{3, t}, F_{4, t}\}$. We drop the subscript of time $t$ to avoid clutter of notation. Here, $e_{p} \in \mathbb{R}^{3}$ is the position error, $e_v \in \mathbb{R}^{3}$ is the velocity error, $e_{\omega} \in \mathbb{R}^{3}$ is the error in body rates, and $R$ is the flattened $3 \times 3$ rotational matrix ($R$ is converted to a $9$-dimensional vector by flattening). Thus, the input state vector to the policy $\pi$ and the adversary $\bar{\pi}$ is of 18 dimensions. For the quadcopter system under consideration, the action space $a \in \mathbb{R}^4$ corresponds to the number of thrust motors, and this also forms the 4-dimension output vector for the policy and adversary networks. The output values are centered at the hovering condition of the quadcopter given by the equation \eqref{eq:hovering-condition}.
\begin{equation}
F_h = \frac{mg}{4}
\label{eq:hovering-condition}
\end{equation}
\noindent where $F_h$ is the required thrust force by each rotor for hovering state of the quadcopter also shown by \cite{grasp}. We scale the policy (and adversary) network outputs from $[-1, 1]$ to the corresponding actuator limits before applying it to the system (a $\tanh$ activation generates outputs in [-1, 1] - see Fig. \ref{fig:network-arch}-(b)). A linear mapping is used to convert neural outputs to quadcopter actuator thrusts given by \eqref{eq:action-scaling} [\cite{adi2020devrlquad}].
\begin{equation}
F_{i} = F_h + \frac{ a_{i} (F_{max} - F_{min})}{2}
\label{eq:action-scaling}
\end{equation}
\noindent where $F_{min}$ and $F_{max}$ are the minimum and maximum thrust forces possible for the actuators respectively, and $i \in \{1,2,3,4\}$ represents the thrust actuator index.

\subsection{Reward Function}
\label{section-reward}
\vspace{-5pt}
We train the drone to reach a fixed waypoint. This waypoint is defined in the world coordinate frame. The desired linear velocity, attitude and body rates are zero at the goal location. The reward earned by the system during each time step is given by equation \eqref{eq:reward-func}.
\begin{equation}
r_{t}= \beta - \alpha_{a}\|a\|_2 - \sum_{_{k\in\{p,v,\omega\}}}\alpha_{k} \|e_{k}\|_2 - \sum_{_{j\in\{\xi,\rho\}}}\alpha_{j}\|e_{j}\|_2
\label{eq:reward-func}
\end{equation}
where $\beta \ge 0$ is a reward for staying alive and coefficients $\alpha_{\{\cdot\}} \ge 0$ represent the weights of various terms in \eqref{eq:reward-func}. The second term in \eqref{eq:reward-func} represents the penalty imposed to avoid excessive actions, followed by the terms representing the cost of error in the quadcopter state. The third term is based on the position error $e_p$, the velocity error $e_v$, and the error in body rates $e_\omega$. The errors in roll and pitch angles, given by $e_\xi$ and $e_\rho$, respectively, are represented in the last term. The RL agent was not penalized for yaw error to prioritize the learning of waypoint navigation irrespective of the drone heading.

\begin{figure*}[ht!]
	\centering
	\includegraphics[width=0.55\textwidth]{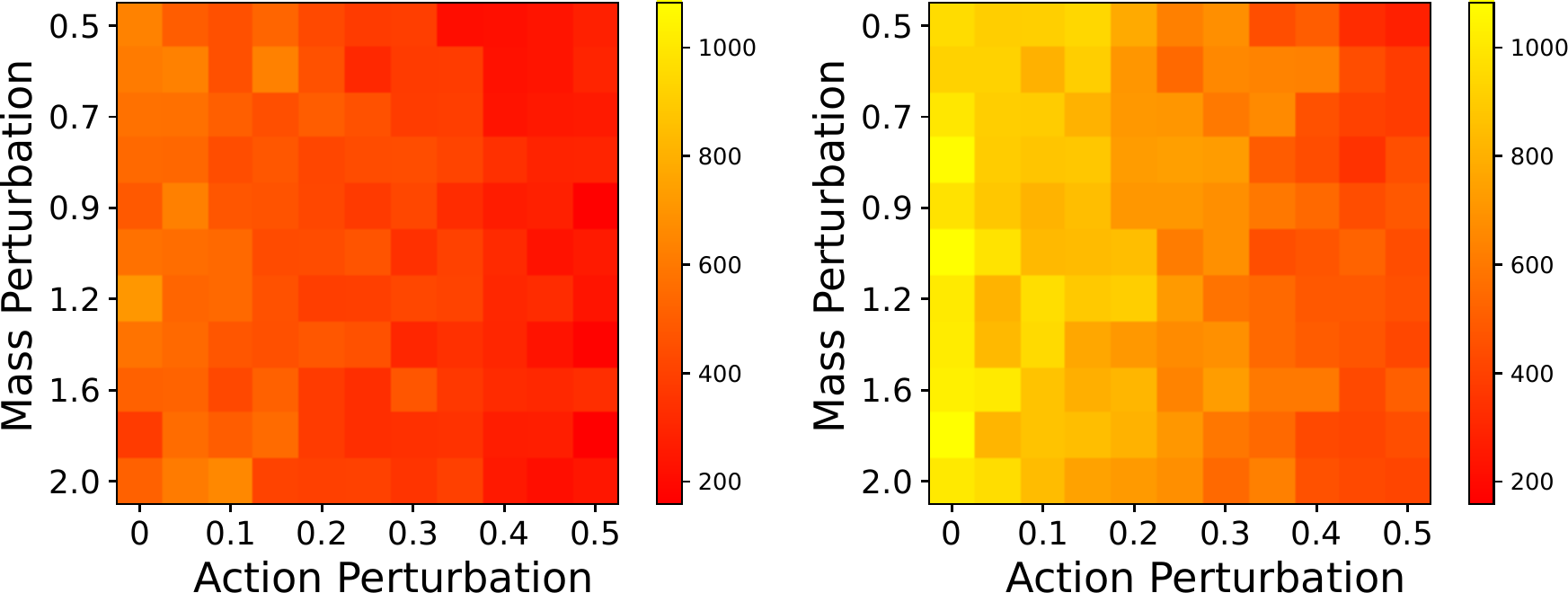}
    \caption{Comparison of policies trained using DDPG (left) and AR-DDPG (right); the color maps represent the average episode return over $10$ episodes for corresponding tuple of (Mass Perturbation, Action Perturbation).}
    \label{fig:results}
\end{figure*}

\subsection{Training Strategy}
\label{sec:training-strategy}
\vspace{-5pt}
We train the control policy to maximize the expected return for the waypoint navigation task. The initial state of the system was sampled randomly from uniform distribution $p_{s_{0}}$. This enables learning the control policy independent of the initial state of the quadcopter. The sampling strategy described by \cite{adi2020devrlquad} was used to initialize the environment for each episode during the training. The drone was trained to reach and hover at fixed waypoint $(0, 0, 5m)$ in the world frame. The initial position $(x, y, z)$ of the quadcopter was sampled uniformly from the cube spanning $2m$ around this location. Initial magnitudes of linear velocity (in $m/s$) and angular velocity (in $rad/s$) were sampled from a uniform distribution in the range of $[0, 1]$. The quadcopter may not be able to recover from certain initial states such as a completely upside down initial orientation, so the uniform sampling of UAV orientation from $SO(3)$ makes the policy training difficult and convergence took longer. Thus, roll, pitch and yaw angles were drawn uniformly from bounded range $[-\frac{\pi}{3}, \frac{\pi}{3}]$. 
The quadcopter desired orientation, linear velocities and angular velocities were kept zero for the task of waypoint navigation.
Each episode in training was terminated if its length exceeded the maximum number of time-steps set in the simulation. Additionally, the episode was terminated if the robot went beyond the bounds defined in the simulation. These bounds were defined as the cube of $2m$ around its target position.

\section{Environment Setup}
\label{sec:env-setup}
\vspace{-5pt}
We used an RL environment as developed by \cite{adi2020devrlquad} for training the control policy. The environment was built using the Mujoco physics engine [\cite{todorov2012mujoco}] and OpenAI Gym [\cite{gym2016openai}]. Table-\ref{table:quadcopter_parameters} provides the physical parameters used to simulate this robot. The coefficients in equations \eqref{eq:reward-func} were: $\alpha_{p}\!\!=\!\!1.0$, $\alpha_{v}\!\!=\!\!0.05$, $\alpha_{\omega}\!\!=\!\!0.001$, $(\alpha_{\xi}, \alpha_{\rho})\!\!=\!\!(0.02, 0.02)$, $\beta\!\!=\!\!2.0$, $\alpha_{{a}}\!\!=\!\!0.025$.
The simulations were performed on a machine with an Intel-i7 processor, 16GB RAM, and NVIDIA RTX 2070.
The training of the policy developed using Pytorch\footnote{\url{https://pytorch.org/}} took approximately $2$ hours. The values of hyperparameters used in AR-DDPG are provided in Table-\ref{table:hyperparameter_table}. 
We used the Adam optimizer [\cite{kingma2014adam}] to optimize neural networks.

\begin{table}[ht!]
    \centering
    \caption{Quadcopter physics model parameters \label{table:quadcopter_parameters}}
    \begin{tabular}{||c|l||} 
         \hline
         Parameter & Value \\ [0.5ex] 
         \hline
         Timestep & 0.01 sec \\ [0.5ex]
         Motor lag & 0.001 sec \\ [0.5ex]
         Mass $m$ & 1.5 kg \\ [0.5ex]
         Length $2l$ & 0.26 m \\ [0.5ex]
         Motor thrust range $[F_{min}, F_{max}]$ & $[0, 15.0]$ N \\ [0.5ex]
         \hline
    \end{tabular}
\end{table}

\begin{table}[ht!]
    \centering
    \caption{AR-DDPG Hyperparameters \label{table:hyperparameter_table}}
    \begin{tabular}{||c|l||}
         \hline 
         Parameter & Value \\ [0.5ex] 
         \hline
         Training iterations & $2 \times 10^{6}$ 
         \\ [0.5ex]
         Learning rate $\eta_{\pi, \bar{\pi}}$ for policy & $2e-5$ \\
         and adversary networks   & \\ [0.5ex] 
         Learning rate $\eta_{Q}$ for critic network & $2e-4$ 
         \\ [0.5ex] 
         Neural network optimizer & Adam 
         \\ [0.5ex] 
         Adam parameters $(\beta_1, \beta_2)$ & $(0.9, 0.999)$ 
         \\ [0.5ex] 
         Discount factor $\gamma$ & $0.95$ 
         \\ [0.5ex] 
         Max. episode length $T$ & $1500$ 
         \\  [0.5ex] 
         Batch size $\lvert B \rvert$ & $64$ 
         \\  [0.5ex]
         Adversary action probability $\alpha$ & $0.1$
         \\ [0.5ex]
         Replay buffer size $\lvert \mathcal{R} \rvert$ & $8 \times 10^{5}$ 
         \\ [0.5ex]
         Policy training steps $n$ & $20$ 
         \\ [0.5ex]
         \hline
    \end{tabular}
\end{table}

\section{Results}
\label{sec:results}
\vspace{-5pt}
We trained two sets of policies, one using robust RL with AR-DDPG and the other non-robust RL using DDPG. In the case of plain DDPG, no adversary was used and the other parameters were kept the same as shown in Table-\ref{table:hyperparameter_table}. 
Although $0 \le \alpha \le 1$, we chose $\alpha\!\!=\!\!0.1$. For higher values of $\alpha$, the adversary gets more control and training a policy becomes difficult. This was also reported by \cite{tessler2019action}. To check the robustness of the control policy trained using the proposed method of AR-DDPG for the UAV, we measured the performance of the drone for waypoint navigation using the reward function \eqref{eq:reward-func} to evaluate the cumulative reward per episode. While testing this policy, both the external and internal parameters to the quadcopter were changed while the trained policy weights were kept unchanged. We did not use the adversary network or the critic network in the testing phase.

We chose the drone mass as the internal parameter and perturbation noise added to the actions as the external parameter to the system. The change in mass simulates the scenarios of the quadcopter system identification errors. We modify the mass of the system $core$ shown in Fig. \ref{fig:quadrotor-x-mujoco} while testing our policies. The appropriate value was added or subtracted from the mass of the $core$ while testing. The external perturbations inserted in the actions can be viewed as the effect of wind on the drone. Table-\ref{table:test-parameters} represent the values of the parameters used for testing the policies. Here, the relative mass is the ratio of the quadcopter masses used during testing $m_{test}$ and training $m_{train}$, where the $m_{train}\!\!=\!\!m$ is given in Table-\ref{table:quadcopter_parameters}. 
The external perturbations were sampled from a uniform distribution and included in actions from trained policy as $a_{i} \!\!=\!\! a_{i} \!+\! \mathrm{Uniform}(-1, 1) \text{\;} \forall i\!\in\!\!\{1, 2, 3, 4\}$ with probability $\delta\;$(given in Table-\ref{table:test-parameters}).
It should be noted that the values in this table were not shown to the system during training. 

\begin{table}[ht!]
    \centering
    \caption{Test parameters \label{table:test-parameters}}
    \begin{tabular}{||c|l||} 
         \hline
         Parameter & Set of Values \\ [0.5ex] 
         \hline
         Quadcopter relative mass $\frac{m_{test}}{m_{train}}$ & $\{0.5, 0.6, 0.7, 0.8, 0.9, 1.0, 1.2,$ \\
         (Mass Perturbations) & $ \quad 1.4, 1.6, 1.8, 2.0\}$
         \\ [1.0ex]
         Probability of external perturbations $\delta$ & $\{ 0, 0.05, 0.1, 0.15, 0.2, 0.25, $ \\
          (Action Perturbations) & $ \quad 0.3, 0.35, 0.4, 0.45, 0.5 \}$
         \\ [0.5ex]
         \hline
    \end{tabular}
\end{table}

Figure \ref{fig:results} contains the color-maps of the results representing the averaged episode returns over all test runs of the trained policies. For each episode ${i}$, the episode return is calculated as $R_{i} \!\!=\!\! \sum_{t=0}^{t_{final}} r_{t,i}$ and $r_{t, i}$ is obtained using equation \eqref{eq:reward-func}.
For each value of the external and internal parameters, the parameters of the policies trained using AR-DDPG and DDPG were kept frozen. The drone was then initialized at a state sampled using the strategy described in Section-\ref{sec:training-strategy} and allowed to perform the task of reaching the goal location. This test was performed $10$ times for each pair of internal and external parameters.
It can be observed from these results that the robust policy trained using AR-DDPG acquires higher rewards compared to the DDPG policy when the environment parameters are varied from the parameters seen by the policy during training. This clearly shows the robustness of the trained neural policy to unseen environmental factors. The robust policy was able to perform comparatively better than the policy trained using DDPG even when no internal or external perturbations are introduced in the environment. Another interesting point to note from these results is that the training of robust policy did not require any knowledge about the model uncertainties in set $\mathcal{P}$ except the scalar value $\alpha$. The adversary $\bar{\pi}$, trained for taking the worst possible action, allowed the controller to be implicitly prepared for the range of uncertainties that may arise in the drone model.

\section{Conclusion and Future Work}
\label{sec:conclusions}
\vspace{-5pt}
In this work, we investigated robust deep reinforcement learning for training control policies for quadcopters. We compared the robust control policy trained using AR-DDPG with the policy trained with the DDPG algorithm. The results suggest that the learned robust policies are able to account for uncertainties in the drone model that are not considered at training time. The results also show that the policy is able to perform better compared to the non-robust policy even in the absence of noise and achieve a higher reward. The drone trained with this approach would also be tolerant to external disturbances in the environment such as wind. Although the value of $\alpha$ implicitly considers uncertainty set for the model, the robust policy may not be able to handle any kind of uncertainty. This is also observable from the results in Fig. \ref{fig:results} where the increase in uncertainty degraded robust policy performance. Thus, depending on the application, one may consider domain randomization as a complement to robust RL.

The approach of using adversarial agents in reinforcement learning could be helpful to mitigate the gap in simulation to real-world transfer of learning-based approaches used to train robots. Robust control policies could account for model uncertainties making the simulation-to-real-world transfer of agents implicitly possible [\cite{loquercio2019deep}]. This work only investigated the symbolic state space for the drones. It would be interesting to see the use of this approach in applications such as intelligent visual servoing using aerial vehicles [\cite{bonatti2020learning}] that would consist of high dimensional raw sensor data such as images.

\vspace{-4pt}
\bibliography{bibliography} % 
\end{document}